\documentclass[11pt]{article}

\usepackage[preprint]{acl}

\usepackage{times}
\usepackage{latexsym}
\usepackage[T1]{fontenc}
\usepackage[utf8]{inputenc}
\usepackage{microtype}
\usepackage{inconsolata}
\usepackage{graphicx}

\usepackage{amsmath}
\usepackage{amssymb}
\usepackage{array}
\usepackage{booktabs}
\usepackage{multirow}
\usepackage{tabularx}

\title{UniEvo-RS: Omni-Prompt Unified Remote Sensing Segmentation with Representative Exemplar-Driven Prototype Evolution}

\author{
  \textbf{Kunquan Zhang\textsuperscript{1}},
  \textbf{Peilang Li\textsuperscript{1}},
  \textbf{Xikun Hu\textsuperscript{2}}
  \\
  \textbf{Yunkai Yang\textsuperscript{1}},
  \textbf{Yushan Zou\textsuperscript{2}},
  \textbf{Zhiwei Zhang\textsuperscript{1}},
  \textbf{Runmin Dong\textsuperscript{1}}
  \\[2mm]
  \textsuperscript{1}Sun Yat-sen University
  \qquad
  \textsuperscript{2}National University of Defense Technology
  \\[1mm]
  \texttt{zhangkq8@mail2.sysu.edu.cn}
  \qquad
  \texttt{dongrm3@mail.sysu.edu.cn}
}

\begin{document}

\maketitle

\begin{abstract}
Prompt-driven vision-language models (VLMs) hold immense promise for accelerating dense remote sensing (RS) annotation, but static models suffer from severe performance degradation when deployed on novel scenes, unseen categories, or visually confusing backgrounds. Moreover, existing unified paradigms primarily rely on intra-image specific prompts, lacking flexible task routing to adapt to multi-intent operational workflows. In practical batch mapping, annotators typically refine a small set of representative samples before processing large datasets. Motivated by this practice, we propose UniEvo-RS, an omni-prompt unified RS segmentation framework equipped with representative exemplar-driven prototype evolution. First, we construct a multi-instruction prompt dataset that unifies text-driven and visual-driven prompts within a single architecture, establishing a dynamic task-routing mechanism for highly diverse RS annotation scenarios. Second, we introduce a representative feedback-driven, training-free prototype evolution mechanism. By contrasting manual annotations with initial predictions on exemplars, UniEvo-RS distills prediction errors into positive and negative prototypes. These prototypes enhance LLM query recall and suppress spatial background noise under a fixed-budget clustering memory. Extensive experiments show that UniEvo-RS unifies diverse prompting tasks, achieving state-of-the-art performance across most settings. Crucially, with minimal interaction on a few exemplars, it enables training-free, progressive accuracy enhancement on unseen categories during batch annotation.
\end{abstract}

\section{Introduction}
\label{sec:introduction}

Semantic segmentation of remote sensing (RS) imagery plays a crucial role in Earth observation tasks such as urban planning and disaster assessment. However, due to the complexity of RS scenes, the highly specialized nature of geographical categories, and the strict requirement for high-fidelity delineation, segmentation models conventionally rely on time-consuming and expensive dense pixel-level annotations~\cite{waqas2019isaid,wang2023samrs}. Recently, prompt-driven vision-language models (VLMs)~\cite{kirillov2023segment,radford2021learning,ravi2024sam,
zou2023generalized,zou2023segment} have demonstrated remarkable zero-shot generalization capabilities in segmentation tasks. By incorporating textual instructions and visual prompts, such as points and bounding boxes, VLMs offer a novel paradigm for constructing initial annotations, enhancing interactive efficiency, and reducing annotation costs~\cite{lai2024lisa,zhang2024psalm,ren2024pixellm}.

\begin{figure*}[t]
    \centering
    \includegraphics[width=0.98\textwidth]{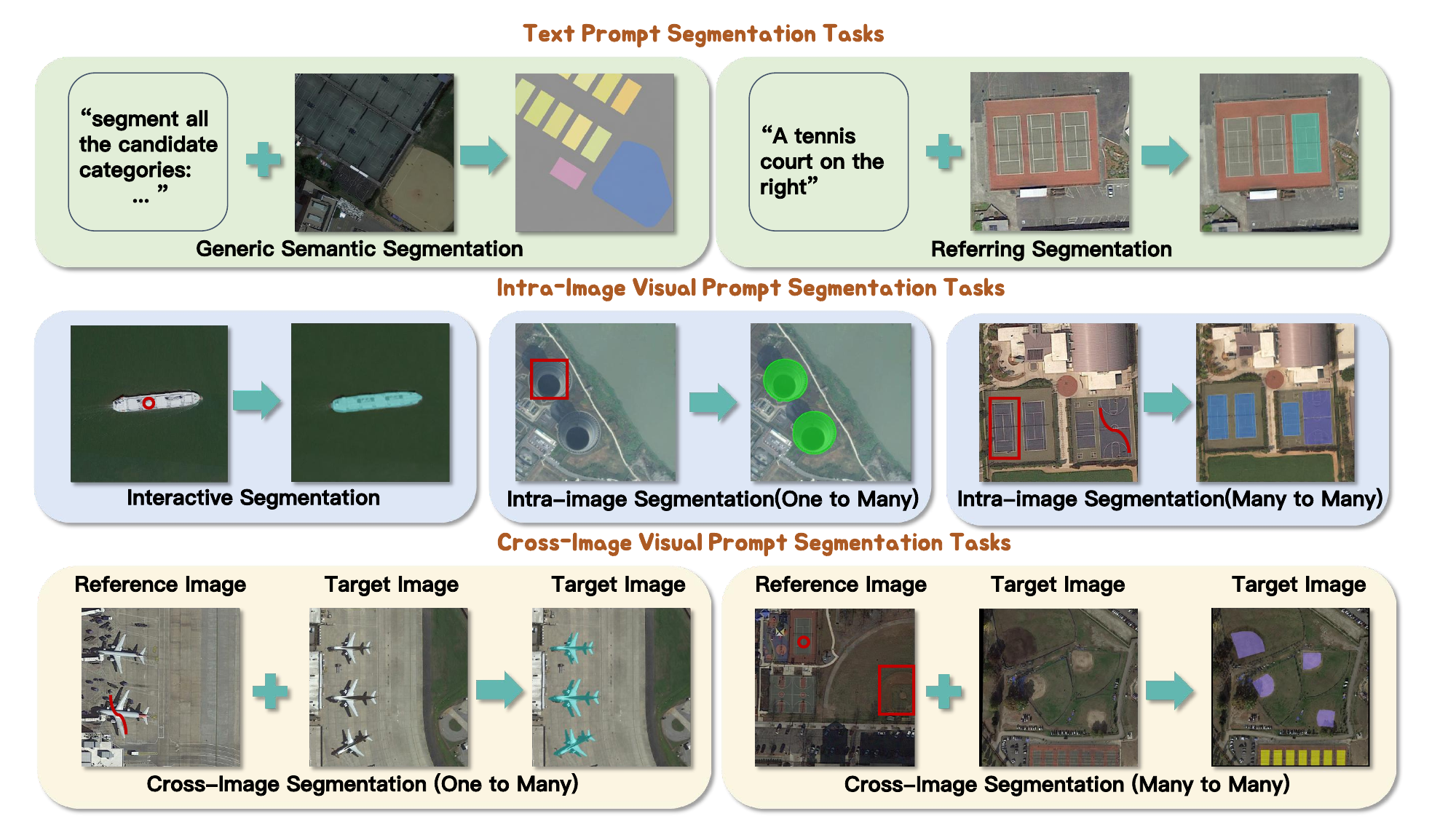}
    \caption{Overview of the segmentation tasks supported by UniEvo-RS. The
    unified architecture integrates text-prompted generic and referring
    segmentation with visual-prompted interactive, intra-image, and cross-image
    segmentation.}
    \label{fig:teaser}
    \vspace{-13pt}
\end{figure*}

Despite these advances, directly applying existing prompt-driven or unified segmentation models to complex RS annotation workflows remains challenging. First, current models primarily rely on intra-image-specific prompts and are typically designed for static, single-turn inference~\cite{kirillov2023segment,ravi2024sam,lai2024lisa,
zhang2024psalm,ren2024pixellm}. Second, they lack flexible task-routing capabilities, making it difficult to switch among the diverse operational intents of human annotators. In practical batch-mapping scenarios, an annotator may need to alternate among referring segmentation, single-instance extraction, and cross-image multi-category segmentation based on a visual exemplar. Existing frameworks do not fully unify and route these heterogeneous instructions within a single architecture, limiting their applicability to comprehensive RS annotation workflows~\cite{wang2023seggpt,li2024visual,zheng2025omni}.

The limitations of static inference become particularly pronounced in practical batch annotation. Although VLMs possess open-vocabulary recognition capabilities, their performance often degrades when applied to RS scenes with novel textures, unseen categories, or confusing backgrounds, resulting in unreliable initial pre-annotations. In realistic batch mapping, human annotators typically inspect and refine a small set of representative samples, or exemplars, before applying the model to a larger image collection. While fine-tuning the model on these verified exemplars is a straightforward solution, repeatedly updating a large VLM introduces prohibitive computational overhead~\cite{lai2024lisa,zhang2024psalm,ren2024pixellm,jia2022visual,
hu2022lora,chen2022adaptformer}. To avoid such repeated parameter updates, methods such as SegGPT~\cite{wang2023seggpt} and OmniRIS~\cite{zheng2025omni} introduce cross-image prompting. However, these methods still process each target image independently and do not explicitly reuse the verified prediction errors obtained from representative exemplars. Directly caching all exemplar features would also introduce unnecessary memory and computational costs. Therefore, extracting reusable target and distractor priors from a small number of human-refined exemplars, while maintaining a fixed memory budget and avoiding parameter updates, remains an important challenge for RS batch annotation.

To address multi-instruction adaptability and batch annotation efficiency, we propose UniEvo-RS, an omni-prompt unified RS segmentation framework featuring representative exemplar-driven prototype evolution. First, we construct a multi-instruction dataset that unifies text and visual prompts. UniEvo-RS maps these heterogeneous prompts into a shared token space, establishing dynamic task routing for diverse annotation intents. Second, we introduce a training-free prototype evolution mechanism. By contrasting manual annotations with initial predictions on exemplars, it distills missed targets and false-positive backgrounds into positive and negative prototypes. Maintained via a fixed-budget clustering strategy, these prototypes enhance query recall and suppress spatial background noise during downstream cross-image inference without parameter updates. Evaluations across multiple benchmarks confirm that the unified architecture achieves state-of-the-art results, and the evolution mechanism yields training-free accuracy improvements on novel distributions. In summary, the main contributions of this paper are as follows:

(1) We propose UniEvo-RS, an omni-prompt unified RS segmentation framework, together with a comprehensive multi-instruction and multimodal-prompt RS segmentation dataset. It establishes a flexible task-routing mechanism that unifies text-prompted and visual-prompted tasks within a single architecture, supporting diverse RS segmentation and annotation intents.

(2) We propose a training-free prototype evolution mechanism for batch annotation. It distills verified errors on exemplars into positive and negative prototypes, which respectively enhance target recall and suppress background noise under a fixed-budget clustering memory.

(3) Extensive experiments demonstrate that UniEvo-RS achieves state-of-the-art results across most of the evaluated tasks. Furthermore, limited exemplar verification significantly enhances training-free cross-image pre-annotation for unseen categories and complex backgrounds.

\section{Related Work}
\label{sec:related_work}

\subsection{Unified Vision-Language Segmentation Models}

Early unified segmentation models sought to accommodate heterogeneous vision tasks within a shared architecture through task-specific prompts or output interfaces~\cite{lu2022unified,cheng2022masked,kolesnikov2022uvim}. The SAM series~\cite{kirillov2023segment,ravi2024sam} further established promptable segmentation using points, boxes, and masks, while SegGPT and Painter~\cite{wang2023seggpt,wang2023images} cast visual tasks as in-context image transformation. X-Decoder and SEEM~\cite{zou2023generalized,zou2023segment} subsequently aligned language and visual conditions in a shared semantic space. More recently, OmniRIS and its baseline OmniSegNet~\cite{zheng2025omni} extended referring segmentation to text and reference-image prompts, including one-to-many and many-to-many settings. Despite their broad task coverage, these models are primarily designed for natural images or intra-image interaction. Their prompt interfaces do not explicitly address the task routing and cross-scene appearance variation required by multi-intent RS annotation. In contrast, UniEvo-RS formulates text-driven and visual-driven RS segmentation as an omni-prompt multi-instruction learning problem, enabling diverse annotation intents to be dynamically routed within a shared architecture.

\subsection{Interactive and Cross-Image RS Segmentation}

Multimodal large language models have enabled segmentation systems to interpret increasingly complex language instructions~\cite{zhu2023minigpt,bai2023qwen}. LISA, PSALM, and PixelLM~\cite{lai2024lisa,zhang2024psalm,ren2024pixellm} connect language-model outputs with pixel-level decoders for referring, reasoning, and interactive segmentation. In remote sensing, GeoChat, EarthMarker, UniGeoSeg, SegEarth-R2, and SkyEyeGPT~\cite{kuckreja2024geochat,zhang2024earthmarker,ni2025unigeoseg,xin2025segearth,zhan2025skyeyegpt} improve geospatial understanding through RS-specific instruction tuning and spatial grounding. However, most of these methods infer each target image from its current prompt and do not explicitly reuse human-verified prediction errors from representative samples. Although in-context models such as SegGPT and OmniRIS enable cross-image visual prompting, they merely treat historical exemplars as simple visual prompts, lacking a deep mining of the underlying information. In contrast, we propose a representative feedback-driven prototype evolution mechanism that deeply decouples verified errors on exemplars into reusable positive and negative prototypes.

\subsection{Representative Feedback-Driven Prototype Adaptation}

Test-time and continual adaptation methods update pretrained models using unlabeled target data or pseudo-labels~\cite{wang2021tent,wang2022cotta,niu2023towards,yuan2023robust}. Although effective under certain distribution shifts, gradient-based adaptation can be computationally expensive for large vision-language models and sensitive to noisy predictions. Parameter-efficient alternatives adapt prompts or feature prototypes instead of the full network~\cite{rebuffi2017icarl,jia2022visual,shu2022test,zhang2022tip}, but they primarily emphasize positive class alignment or adaptation to individual samples. Meanwhile, conventional interactive segmentation typically treats human corrections as immediate prompts without converting verified errors into reusable cross-image knowledge. To bridge this gap, UniEvo-RS introduces a training-free approach. By explicitly storing verified feedback in a fixed-budget prototype memory, it transfers the verified knowledge to new target data through forward-pass inference.


\section{Methodology}
\label{sec:methodology}

\begin{figure*}[t]
    \centering
    \includegraphics[width=0.98\textwidth]{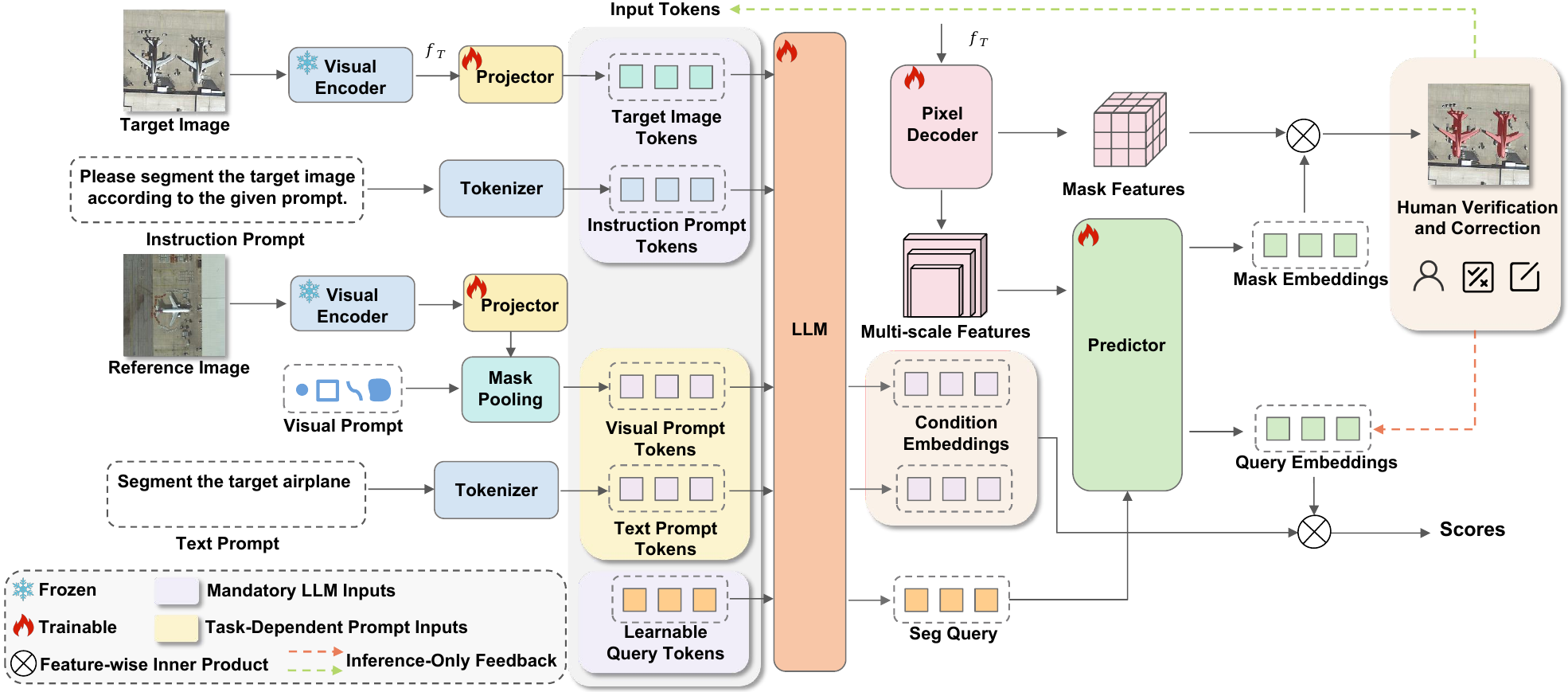}
    \caption{Overall architecture of UniEvo-RS.}
    \label{fig:overall_architecture}
\end{figure*}

\begin{figure*}[t]
    \centering
    \includegraphics[width=0.98\textwidth]{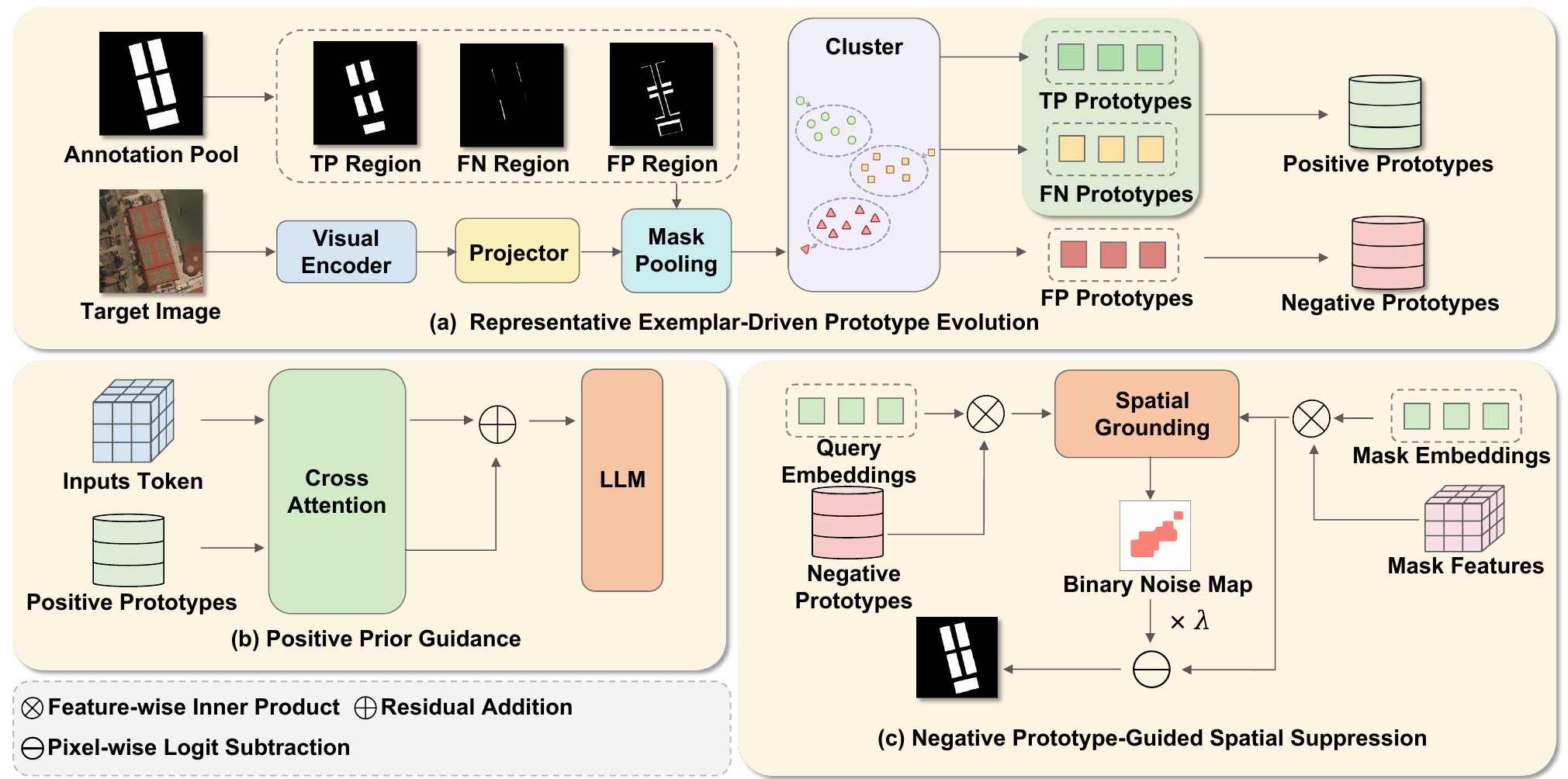}
    \caption{Representative exemplar-driven prototype evolution. Human-verified
    TP/FN regions form positive target memory, whereas FP regions form negative
    background memory. K-Means maintains fixed budgets for subsequent cross-image
    inference without parameter updates.}
    \label{fig:dynamic_prototype_evolution}
    \vspace{-11pt}
\end{figure*}

\subsection{Overall Workflow}
\label{sec:overall_workflow}

To address multi-instruction adaptability and batch annotation efficiency, we propose UniEvo-RS. It is an omni-prompt unified RS segmentation framework equipped with representative exemplar-driven prototype evolution. Our method consists of two primary mechanisms.

First, we design a unified architecture to accommodate diverse annotation intents. As illustrated in Figure~\ref{fig:overall_architecture}, UniEvo-RS processes multi-instruction inputs. Target images, reference features, language instructions, and visual prompts are mapped into a shared token space. An LLM contextualizes these inputs and dynamically routes them according to the current task instruction. A shared mask decoder then produces task-aware masks and matching scores. This formulation seamlessly integrates text-prompted and visual-prompted RS segmentation tasks.

Second, we introduce a training-free prototype evolution mechanism for practical batch annotation. UniEvo-RS compares the manual annotations of representative exemplars with their initial model predictions. It decomposes the prediction errors into missed target regions and incorrectly detected background distractors. Features from missed target regions are clustered into positive prototypes to guide the LLM toward recall-enhanced segmentation queries. False-positive background features form negative prototypes that drive spatial suppression in the mask predictor. A clustering strategy maintains the positive and negative prototype banks under a fixed memory budget. Consequently, the resulting exemplar feedback can be directly reused during downstream cross-image inference without updating any model parameters.

\subsection{Omni-Prompt Unified RS Segmentation Framework}
\label{sec:omni_prompt}

UniEvo-RS projects heterogeneous instructions and multimodal prompts into a shared token space and routes them through a common mask-generation pathway. All tasks are formulated as:

\begin{equation}
\mathbf{Y} = \mathcal{F}(I^{t},\mathcal{I},\mathcal{P}^{l},\mathcal{P}^{v}),
\end{equation}
where $I^{t}$ is the target image, $\mathcal{I}$ is the task instruction, $\mathcal{P}^{l}$ and $\mathcal{P}^{v}$ are optional language and visual prompts, and $\mathbf{Y}$ is the segmentation output.

\paragraph{Visual Feature Extraction via Image Encoder.}
A frozen pretrained vision transformer extracts multi-scale features
$\mathbf{V}^{t}$ and $\mathbf{V}^{p}$ from the target image $I^{t}$ and
prompt image $I^{p}$, respectively. For intra-image prompting,
$I^{t}=I^{p}$; otherwise, the two images are encoded with shared weights.
These features are subsequently utilized for prompt encoding, mask decoding, and prototype
extraction.

\paragraph{Multimodal Prompt Encoding.}
The multimodal prompt encoder maps heterogeneous prompt modalities into a
unified token space. We adopt a shared token interface containing special
tokens such as \texttt{<image>}, \texttt{<region>}, \texttt{<seg>},
\texttt{<cls>}, and \texttt{<refer>}. Text prompts are directly embedded as
sequence tokens. For visual prompts, such as boxes, points, or scribbles, the
encoded prompt-image feature map $\mathbf{V}^{p}$ is first mapped into the LLM
embedding space by a trainable visual projector $g_v$. Mask pooling then
samples and aggregates the projected features within each prompted region:
\begin{equation}
\mathbf{r}_{k}
=
\operatorname{Pool}\!\left(
\operatorname{Sample}\!\left(g_v(\mathbf{V}^{p}),M_{k}^{p}\right)
\right),
\end{equation}
where $M_{k}^{p}$ denotes the $k$-th prompt mask and $\mathbf{r}_k$ is its
visual prompt token. A modality selector routes text tokens, visual tokens,
or both into the LLM according to the task instruction. This design supports
one-to-many and multi-region visual prompting under a unified formulation.

\paragraph{LLM Contextualization.}
The target-image tokens, selected prompt tokens, task instruction, and
learnable segmentation queries are concatenated into a multimodal sequence
$\mathbf{X}$ and fed into the LLM:
\begin{equation}
\mathbf{H}=\mathcal{F}_{\mathrm{llm}}(\mathbf{X}).
\end{equation}
From $\mathbf{H}$, we extract contextualized segmentation queries at the
\texttt{<seg>} positions and condition embeddings at the prompt-token
positions, explicitly encoding task-aware reasoning and target conditions,
respectively.

\paragraph{Unified Mask Decoder.}
The unified mask decoder consists of a pixel decoder and a shared mask
predictor. The pixel decoder transforms the target-image backbone features
into multi-scale decoder features $\mathbf{F}_{\mathrm{multi}}$ and
high-resolution mask features $\mathbf{F}_{\mathrm{mask}}$. The mask predictor
uses the contextualized segmentation queries $\tilde{\mathbf{q}}_q$ to attend
to $\mathbf{F}_{\mathrm{multi}}$, producing decoded query embeddings:
\begin{equation}
\mathbf{z}_{q}
=
\mathcal{D}_{\mathrm{pred}}
\left(\tilde{\mathbf{q}}_{q},\mathbf{F}_{\mathrm{multi}}\right).
\end{equation}

Each decoded query is mapped to a mask embedding by a mask MLP $\phi_m$, and
the query-wise candidate mask logits are computed as
\begin{equation}
\hat{\mathbf{M}}_{q}
=
\left\langle
\phi_m(\mathbf{z}_{q}),
\mathbf{F}_{\mathrm{mask}}
\right\rangle.
\end{equation}

For region-conditioned instructions, the matching score between the $k$-th
condition embedding and the $q$-th decoded query is
\begin{equation}
s_{k,q}
=
\phi_r(\mathbf{z}_{q})^{\top}\mathbf{r}'_{k},
\end{equation}
where $\phi_r$ projects decoded queries into the shared region-query embedding
space and $\mathbf{r}'_k$ is the projected condition embedding. This design
keeps the computationally intensive mask-generation pathway shared across
tasks.

\paragraph{Unified Multi-Task Optimization.}
During training, we adopt a batch-wise task alternation strategy, where each mini-batch is sampled from one task and tasks rotate across batches. All tasks share the mask loss $\mathcal{L}_{\mathrm{mask}}$ and Dice loss $\mathcal{L}_{\mathrm{dice}}$ for mask-quality supervision. Task-specific matching losses $\mathcal{L}_{\mathrm{prompt}}$, such as cross-entropy losses for semantic categories or referring expressions, are dynamically activated according to the instruction type:

\begin{equation}
\mathcal{L}
=
\lambda_{\mathrm{mask}}\mathcal{L}_{\mathrm{mask}}
+
\lambda_{\mathrm{dice}}\mathcal{L}_{\mathrm{dice}}
+
\lambda_{\mathrm{prompt}}\mathcal{L}_{\mathrm{prompt}}.
\end{equation}

At inference time, a lightweight task-specific post-processing step parses the unified predictions. For referring segmentation, the mask with the highest matching score is selected. For interactive and generic segmentation, masks exceeding a predefined confidence threshold are retained, adapting the unified output to different RS task requirements.

\subsection{Representative Exemplar-Driven Prototype Evolution}
\label{sec:prototype_evolution}

Although the unified architecture supports multiple annotation intents, its initial predictions may still degrade on unseen target appearances and confusing backgrounds in novel RS image collections. In practical batch annotation, an annotator can first verify a small set of representative exemplars before applying the model to the remaining images. We therefore introduce representative exemplar-driven, training-free prototype evolution to reuse this verified feedback without parameter updates.

\paragraph{Prototype Memory Construction.}
For each representative exemplar, the model first predicts $\hat{\mathbf{Y}}$, after which an annotator verifies or corrects it to obtain $\mathbf{Y}_{v}$. Their comparison decomposes the result into true positives (TP), false negatives (FN), and false positives (FP). To simulate this interactive verification, ground-truth annotations serve as a surrogate for human feedback, revealed strictly after the initial prediction.
We extract region-level prototype tokens from the projected target-image
feature map $\bar{\mathbf{V}}=g_v(\mathbf{V})$, using the same visual
projector $g_v$ introduced above, where $\mathbf{V}$ denotes the target-image
feature map produced by the image encoder. Specifically, a prototype $\mathbf{r}\in\mathbb{R}^{d}$ is generated through mask pooling:

\begin{equation}
\mathbf{r}
=
\operatorname{Pool}\!\left(
\operatorname{Sample}(\bar{\mathbf{V}},M)
\right).
\end{equation}

where $M$ denotes a decomposed region mask, such as $M_{\mathrm{FN}}$. These tokens are stored in three isolated partitions, $\mathcal{B}_{\mathrm{TP}}$, $\mathcal{B}_{\mathrm{FN}}$, and $\mathcal{B}_{\mathrm{FP}}$. TP and FN prototypes serve as positive memory for target guidance, while FP prototypes serve as negative memory for background suppression.

\paragraph{Representative Exemplar-Driven Prototype Evolution.}
To keep the prototype memory exposed to the model within a fixed budget, we
compress newly verified evidence and the previously retained centers through
K-Means clustering. Let
$\mathcal{R}^{(t)}_{c}$ denote the newly extracted region tokens of type
$c\in\{\mathrm{TP},\mathrm{FN},\mathrm{FP}\}$ at representative step $t$. A
temporary update pool is constructed as
\begin{equation}
\mathcal{U}^{(t)}_{c}
=
\mathcal{M}^{(t-1)}_{c}
\cup
\mathcal{R}^{(t)}_{c}.
\end{equation}

K-Means is applied independently to the TP, FN, and FP update pools.
Whenever newly verified tokens are added, the retained centers are recomputed
from the temporary pools, allowing the memory to incorporate feedback from
successively verified representative exemplars. The positive and negative
memories at step $t$ are defined as
\begin{equation}
\mathcal{M}^{+}_{t}
=
\left[
\mathcal{C}_{m}\!\left(\mathcal{U}^{(t)}_{\mathrm{TP}}\right);
\mathcal{C}_{n}\!\left(\mathcal{U}^{(t)}_{\mathrm{FN}}\right)
\right],
\end{equation}
\begin{equation}
\mathcal{M}^{-}_{t}
=
\mathcal{C}_{r}\!\left(\mathcal{U}^{(t)}_{\mathrm{FP}}\right),
\end{equation}
where $\mathcal{C}_{k}(\cdot)$ extracts at most $k$ cluster centers from the
corresponding temporary pool. After clustering, only the centers are retained
and the temporary pools are discarded. Thus, memory always contains at most
$m+n$ positive prototypes and $r$ negative prototypes. Retaining both TP and
FN centers preserves stable target characteristics and difficult patterns
associated with missed targets. The hyperparameters $m$ and $n$ control the
allocation between stable and difficult target priors.

\paragraph{Positive Prior Guidance.}
Before LLM contextualization, positive memory $\mathcal{M}^{+}$ is injected into the current multimodal sequence $\mathbf{X}$. We compute a memory-enhanced representation $\tilde{\mathbf{X}}$ through residual cross-attention:

\begin{equation}
\mathbf{A}^{+}
=
\operatorname{Softmax}\!\left(
\frac{\mathbf{X}(\mathcal{M}^{+})^{\top}}{\sqrt{d}}
\right),
\end{equation}

\begin{equation}
\tilde{\mathbf{X}}
=
\mathbf{X}
+
\alpha\mathbf{A}^{+}\mathcal{M}^{+},
\end{equation}
where $\alpha$ modulates the injection strength. This operation enriches the input tokens with target characteristics and missed patterns extracted from verified representative exemplars, guiding the LLM queries toward relevant target regions in subsequent images.

\paragraph{Negative Prototype-Guided Spatial Suppression.}
Negative memory $\mathcal{M}^{-}$ is employed after query decoding to suppress spatial patterns associated with false-positive backgrounds identified from verified representative exemplars. Rather than directly generating masks, each negative prototype $\mathbf{p}^{-}_{k}$ and decoded query embedding $\mathbf{z}_{q}$ are projected into a shared region-query space. Their negative matching affinity is computed as
\begin{equation}
a^{-}_{kq}
=
\sigma\left(
\frac{\left\langle\phi_n(\mathbf{p}^{-}_{k}),\phi_r(\mathbf{z}_{q})\right\rangle}{\sqrt{d_r}}
\right),
\end{equation}
where $\phi_n$ and $\phi_r$ are projection functions, $d_r$ is the matching dimension, $\sigma(\cdot)$ is the sigmoid function, and $k$ and $q$ index the negative prototypes and decoded queries, respectively.

For each decoded query, we retain its strongest affinity with the negative memory:
\begin{equation}
w^{-}_{q}
=
\max_{k} a^{-}_{kq}.
\end{equation}

The resulting confidence $w^{-}_{q}$ measures how strongly the $q$-th query resembles any verified false-positive prototype.

Let $M_q(u,v)$ denote the candidate mask logit predicted by query $q$ at spatial location $(u,v)$. We spatially ground the query-level negative confidence onto the corresponding candidate mask and aggregate the responses over all decoded queries:
\begin{equation}
S^{-}(u,v)
=
\max_{q}
\left[
w^{-}_{q}
\,
\sigma\left(M_q(u,v)\right)
\right],
\end{equation}
where $S^{-}(u,v)$ denotes the negative spatial confidence map. This operation ensures that a candidate mask contributes strongly to the negative spatial response only when its corresponding query is highly similar to at least one negative prototype.

A binary noise map is then obtained through confidence thresholding:
\begin{equation}
N(u,v)
=
\mathbb{I}\left(
S^{-}(u,v)>\tau
\right),
\end{equation}
where $\mathbb{I}(\cdot)$ denotes the indicator function and $\tau$ is the confidence threshold.

Finally, the candidate mask logits are refined through explicit spatial subtraction:
\begin{equation}
\widetilde{M}_{q}(u,v)
=
M_q(u,v)
-
\lambda N(u,v),
\end{equation}
where $\lambda$ controls the spatial suppression strength. The same noise map is applied to all candidate masks, suppressing regions jointly supported by verified false-positive prototypes and their matched queries. This operation modifies only the spatial mask logits while leaving condition-query matching scores unchanged. Both guidance paths are used only during representative exemplar-driven batch inference, with all model parameters fixed.

\begin{table}[t]
\centering
{\small
\setlength{\tabcolsep}{1.2pt}
\renewcommand{\arraystretch}{1.08}
\begin{tabular*}{\columnwidth}{@{\extracolsep{\fill}}lcccccc@{}}
\toprule
\multirow{2}{*}{\textbf{Method}} & \multicolumn{2}{c}{\textbf{Interactive}} & \multicolumn{2}{c}{\textbf{Intra 1:N}} & \multicolumn{2}{c}{\textbf{Cross 1:N}} \\
\cmidrule(lr){2-3}\cmidrule(lr){4-5}\cmidrule(lr){6-7}
& \textbf{gIoU} & \textbf{cIoU} & \textbf{gIoU} & \textbf{cIoU} & \textbf{gIoU} & \textbf{cIoU} \\
\midrule
SEEM & 15.63 & 3.80 & -- & -- & -- & -- \\
PSALM & 17.10 & 9.36  & -- & -- & -- & -- \\
X-SAM & 18.99 & 6.60 & -- & --& -- & -- \\
UniGeoSeg & 30.41 & 38.96 & -- & -- & -- & -- \\
SAM3 & 61.60 & 56.55 & -- & -- & 38.16 & 37.71 \\
DINOv & 36.49 & 29.78 & 38.66 & 31.55 & 30.35 & 20.43 \\
YOLOE & 43.10 & 35.40 & 45.46 & 49.40 & 13.75 & 13.86 \\
COSINE & 10.81 & 12.35 & 17.04 & 16.06 & 32.27 & 31.21 \\

\midrule
\multicolumn{7}{l}{\textit{Fine-tuned on our dataset}} \\
\midrule
UniGeoSeg & 54.23 & 59.34 & -- & -- & -- & -- \\
SEEM & 24.75 & 12.99 & -- & -- & -- & -- \\
SAM3 & \textbf{67.57} & \underline{68.54} & -- & -- & -- & -- \\
OmniSegNet & -- & -- & 48.10 & 47.50 & 41.58 & 38.68 \\
YOLOE & 47.27 & 41.57 & 45.00 & 41.94 & 12.58 & 9.78 \\
DINOv & 36.47 & 35.37 & \underline{61.30} & \underline{58.98} & \textbf{74.97} & \underline{74.78} \\
UniEvo-RS (Ours)& \underline{64.71} & \textbf{78.75} & \textbf{66.37} & \textbf{65.53} & \underline{74.49} & \textbf{79.25} \\

\bottomrule
\end{tabular*}
}
\caption{Comparison on visual-prompted segmentation tasks. The upper and lower
blocks report zero-shot and fine-tuned results, respectively. We report gIoU
and cIoU for interactive, intra-image 1:N, and cross-image 1:N segmentation.
A dash indicates that the corresponding task is not supported. Best and
second-best results within each block are shown in bold and underlined,
respectively.}
\label{tab:visual_prompt_results}
\vspace{-14pt}
\end{table}

\begin{table}[t]
\centering
{\small
\renewcommand{\arraystretch}{1.1}
\begin{tabular*}{\columnwidth}{@{\extracolsep{\fill}}lccc@{}}
\toprule
\multirow{2}{*}{\textbf{Method}} & \textbf{Generic} & \multicolumn{2}{c}{\textbf{Referring}} \\
\cmidrule(lr){2-2}\cmidrule(lr){3-4}
& \textbf{PQ} & \textbf{gIoU} & \textbf{cIoU} \\
\midrule
UniGeoSeg & -- & 61.08 & 65.70 \\
SAM3 & -- & 42.57 & 28.65 \\
DINOv & 19.16 & -- & -- \\
SEEM & 26.57 & 20.35 & 18.36 \\
PSALM & 12.55 & 24.75 & 20.84 \\
RemoteSAM & 25.67 & \textbf{66.67} & 42.81 \\
COSINE & 3.29 & 19.93 & 13.07 \\
X-SAM & 10.56 & 32.71 & 22.86 \\
\midrule
\multicolumn{4}{l}{\textit{Fine-tuned on our dataset}} \\
\midrule
SAM3 & -- & 47.10 & 49.37  \\
UniGeoSeg & -- & 63.72 & \textbf{80.06} \\
OmniSegNet & -- & 34.09 & 48.08 \\
SegEarth-R2 & -- & 34.99 & 37.94 \\
DINOv & \underline{42.81} & -- & -- \\
SEEM & 27.96 & 46.59 & 57.47\\
RemoteSAM & 10.74 & \underline{65.75} & \underline{78.96} \\
COSINE & 2.85 & 63.31 & 76.34 \\
UniEvo-RS (Ours) & \textbf{62.60} & 55.72 & 69.83 \\
\bottomrule
\end{tabular*}
}
\caption{Comparison on text-prompted segmentation tasks. The upper and lower
blocks report zero-shot and fine-tuned results, respectively. PQ is used for
generic segmentation, while gIoU and cIoU are used for referring segmentation.
A dash indicates that the corresponding task is not supported. Best results
within each block are shown in bold.}
\label{tab:text_prompt_results}
\vspace{-8pt}
\end{table}

\section{Experiments}
\label{sec:experiments}

\subsection{Datasets and Metrics}
\label{sec:datasets_metrics}

We construct a multi-instruction RS segmentation dataset from four public
benchmarks: SIOR~\citep{wang2023samrs},
iSAID~\cite{waqas2019isaid},
NWPU-Refer~\cite{yang2025large}, and
RRSIS-D~\cite{liu2024rotated}. This dataset encompasses text-prompted generic and
referring segmentation, as well as visual-prompted interactive, intra-image
one-to-many, and cross-image one-to-many segmentation. Dataset statistics,
task construction, and prompt generation protocols are provided in the
appendix.

We report PQ for generic segmentation, and gIoU and cIoU for referring, interactive, and one-to-many visual prompt segmentation. For the prototype-memory ablations, we additionally report mIoU to jointly measure foreground recovery and background suppression.

\subsection{Comparison Results}
\label{sec:comparison_results}
Tables \ref{tab:visual_prompt_results} and \ref{tab:text_prompt_results} benchmark UniEvo-RS against general-purpose segmentation models, including SEEM~\citep{zou2023segment},
PSALM~\citep{zhang2024psalm}, SAM3~\citep{carion2025sam},
DINOv~\citep{li2024visual}, YOLOE~\citep{wang2025yoloe}, and
X-SAM~\citep{wang2026x}, as well as RS-specific methods such as UniGeoSeg~\citep{ni2025unigeoseg}, RemoteSAM~\citep{yao2025remotesam}, and SegEarth-R2~\citep{xin2025segearth}. We further include
OmniSegNet~\citep{zheng2025omni} for the one-to-many settings. The upper blocks evaluate released
checkpoints, whereas the lower blocks report applicable baselines fine-tuned on our dataset's training splits. All methods are evaluated under identical test splits and metrics.

As shown, fine-tuning generally improves applicable baselines. Crucially, UniEvo-RS supports all five settings and achieves state-of-the-art results on five of the nine metrics, including generic segmentation, intra-image one-to-many segmentation, and cIoU for interactive and cross-image one-to-many segmentation. While specialized RS models edge out UniEvo-RS in referring segmentation due to their heavily optimized task-specific language grounding, our framework remains best or highly competitive on most metrics, providing complete task coverage without sacrificing multi-task versatility.

\begin{table}[t]
\centering
\small
\setlength{\tabcolsep}{3.5pt}
\renewcommand{\arraystretch}{1.12}
\begin{tabularx}{\columnwidth}{c>{\centering\arraybackslash}Xcc}
\toprule
\textbf{Exp.} & \textbf{Method} & \textbf{gIoU} & \textbf{mIoU} \\
\midrule
1 & Baseline & 36.43 & 54.96 \\
2 & Previous-mask transfer & 36.66 & 56.87 \\
3 & Positive memory concatenation & 36.47 & 54.92 \\
4 & Positive prior guidance & \textbf{41.36} & 57.39 \\
5 & Positive and negative prior guidance & 41.24 & \textbf{59.50} \\
\bottomrule
\end{tabularx}
\caption{Ablation results of fixed-budget memory fusion on cross-image one-to-many segmentation.}
\label{tab:ablation_main}
\vspace{-20pt}
\end{table}

\subsection{Ablation Study}
\label{sec:ablation_study}

We evaluate the proposed training-free prototype evolution mechanism in a practical batch annotation setup using the Vehicle category from SIOR. Specifically, prototypes are extracted from human-verified feedback on a small set of representative exemplars and subsequently applied to the remaining target batch. As shown in Table~\ref{tab:ablation_main}, transferring the previous
mask or directly concatenating historical features provides limited benefit,
indicating that feedback reuse alone is insufficient. Positive prior guidance
improves target recognition, while incorporating negative guidance further
improves mIoU with comparable gIoU. This behavior is consistent with the
intended complementary roles of positive prototypes in recovering targets and
negative prototypes in suppressing recurring false positives.Additional
analyses of prototype allocation, hyperparameter sensitivity, fixed-budget
efficiency, and memory compression are provided in the appendix.

\subsection{Limitations}
Although UniEvo-RS demonstrates strong performance across multiple remote sensing benchmarks, our current evaluation remains constrained to the considered datasets and annotation configurations, leaving broader cross-sensor and cross-domain generalization for future work. Furthermore, many-to-many segmentation is primarily assessed through qualitative visualizations, as a standardized quantitative benchmark split is currently lacking.


\section{Conclusion}
\label{sec:conclusion}
In this paper, we present UniEvo-RS, an omni-prompt unified framework supporting five text-prompted and visual-prompted RS segmentation tasks, backed by a comprehensive multi-instruction dataset. Tailored for RS batch annotation, UniEvo-RS introduces a training-free prototype evolution mechanism driven by representative feedback. By contrasting manual annotations with initial predictions on a few exemplars, it decomposes errors into missed targets and false-positive distractors, forming fixed-budget positive and negative prototype memories. These memories guide LLM queries and suppress recurring background noise during cross-image inference without updating model parameters. Experiments demonstrate unified performance and confirm that limited human verification on representative exemplars substantially enhances downstream batch pre-annotation.




\bibliography{aaai2027}

\clearpage
\appendix

\section{Appendix}
\label{sec:appendix}

\subsection{Omni-Prompt Unified Remote Sensing Segmentation}
\label{sec:supp_omniprompt}

\subsubsection{Omni-Prompt Multi-Instruction Dataset}
\label{sec:supp_dataset}

To support diverse remote sensing annotation intents within a single
architecture, we reformat task-specific splits derived from SIOR, iSAID,
NWPU-Refer, and RRSIS-D into a unified instruction-following segmentation
dataset. Each constructed sample contains a target image, a task instruction,
an optional textual or visual prompt, and the corresponding segmentation
target. For cross-image one-to-many segmentation, the visual prompt is
provided by a region in a separate source image.

The task instruction is present for all samples and specifies the current
annotation intent. The optional prompt determines whether additional textual
prompt tokens or visual prompt tokens are activated. This unified sample
representation allows heterogeneous segmentation tasks to share the same
multimodal token space, LLM contextualization pathway, and mask-generation
interface while being dynamically routed according to the task instruction.

As summarized in Table~\ref{tab:supp_dataset_composition}, the reconstructed
dataset contains two text-prompted settings and three visual-prompted
settings. The reported train and test statistics denote the numbers of
constructed prompt--target pairs rather than unique images.

\begin{table*}[t]
\centering
\small
\setlength{\tabcolsep}{5pt}
\renewcommand{\arraystretch}{1.12}
\begin{tabular*}{\textwidth}{
@{\extracolsep{\fill}}
llllrr
@{}
}
\toprule
\textbf{Prompt Modality}
& \textbf{Task Setting}
& \textbf{Prompt Form}
& \textbf{Source Dataset(s)}
& \textbf{Train}
& \textbf{Test} \\
\midrule

\multirow{2}{*}{Text}
& Generic
& Semantic task instruction
& SIOR
& 6,034
& 1,492 \\

& Referring
& Natural-language expression
& NWPU-Refer, RRSIS-D
& 23,698
& 2,639 \\

\midrule

\multirow{3}{*}{Visual}
& Interactive
& Point, box, scribble, mask
& SIOR, iSAID
& 44,374
& 1,757 \\

& Intra-image 1:N
& Point, box, scribble, mask
& SIOR
& 67,269
& 19,011 \\

& Cross-image 1:N
& Point, box, scribble, mask
& SIOR
& 67,269
& 19,291 \\

\bottomrule
\end{tabular*}
\caption{Composition of the reconstructed omni-prompt remote sensing
segmentation dataset. Train and test statistics denote the numbers of
constructed prompt--target pairs rather than unique images.}
\label{tab:supp_dataset_composition}
\end{table*}

\subsubsection{Text-Prompted Task Construction}
\label{sec:supp_text_tasks}

\paragraph{Generic semantic segmentation.}
We reorganize SIOR into instruction--mask pairs. The language instruction
specifies the semantic segmentation intent, while the dense semantic
annotations of the target image provide pixel-level supervision. This
setting evaluates whether the shared architecture can perform category-level
dense prediction from a task-level language instruction without requiring
an additional spatial prompt.

\paragraph{Referring segmentation.}
We use NWPU-Refer and RRSIS-D, which associate natural-language referring
expressions with spatially corresponding object masks. Each expression is
used as the textual prompt, and the associated mask is used as the
segmentation target. This setting evaluates whether the model can jointly
resolve linguistic attributes, object identity, and spatial references in
complex remote sensing scenes.

\subsubsection{Visual-Prompted Task Construction}
\label{sec:supp_visual_tasks}

\paragraph{Interactive segmentation.}
We construct visual prompts from annotated foreground regions in SIOR and
iSAID. Points, bounding boxes, scribbles, and region masks are used as
alternative prompt forms, while the mask of the prompted instance provides
supervision. These prompt types expose the model to different levels of
spatial specificity and support conventional single-instance interaction.

\paragraph{Intra-image one-to-many segmentation.}
Given a visual prompt sampled from one instance in the target image, the
model is required to segment all instances belonging to the same semantic
category in that image. This task extends conventional single-instance
interaction to category-level extraction within a single scene.

\paragraph{Cross-image one-to-many segmentation.}
A visual prompt is sampled from a source image and paired with a different
target image containing instances of the same semantic category. The
segmentation target contains all matching instances in the target image.
This task evaluates whether region-level semantics can be transferred across
images despite variations in appearance, scale, orientation, spatial layout,
object density, and background context.

\subsubsection{Qualitative Results for One-to-Many Segmentation}
\label{sec:supp_one_to_many}

We provide additional qualitative comparisons for the two one-to-many
visual-prompted settings. In each example, the region enclosed by the red
bounding box serves as the reference prompt. The model is required to
identify and segment all instances sharing the prompted semantic category.
Green overlays denote the target ground-truth or predicted masks.

\begin{figure*}[t]
    \centering
    \includegraphics[width=0.98\textwidth]{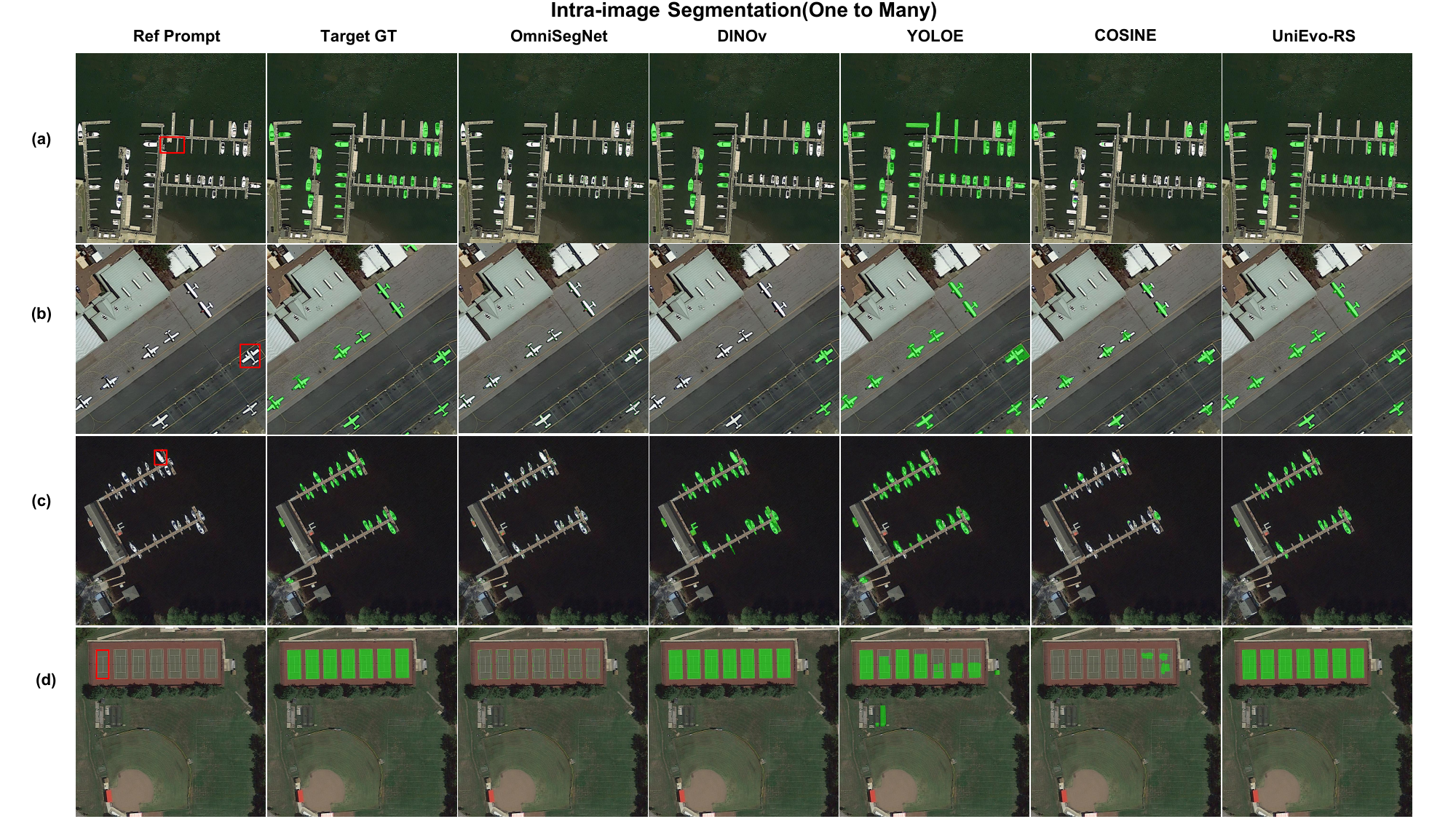}
    \caption{Qualitative comparisons on intra-image one-to-many
    segmentation. The reference prompt and target instances are located
    in the same image. The red bounding box specifies one reference
    instance, and the model is required to segment all instances
    belonging to the same semantic category. Green overlays denote the
    target ground-truth or predicted masks.}
    \label{fig:intra_image_qualitative}
\end{figure*}

Figure~\ref{fig:intra_image_qualitative} illustrates the intra-image
one-to-many setting. In the shown examples, OmniSegNet fails to propagate
the prompted category semantics to several corresponding instances. DINOv
and YOLOE recover more targets but still miss some small, densely distributed,
or appearance-shifted instances. COSINE occasionally produces incomplete
masks or responses on visually similar background structures.

Across the displayed cases, UniEvo-RS produces more complete target coverage
and fewer fragmented responses. The improvement is particularly visible for
densely distributed aircraft and ships, as well as repeated structures such
as sports courts, where instances exhibit substantial variation in scale,
orientation, and surrounding context.

\begin{figure*}[t]
    \centering
    \includegraphics[width=0.98\textwidth]{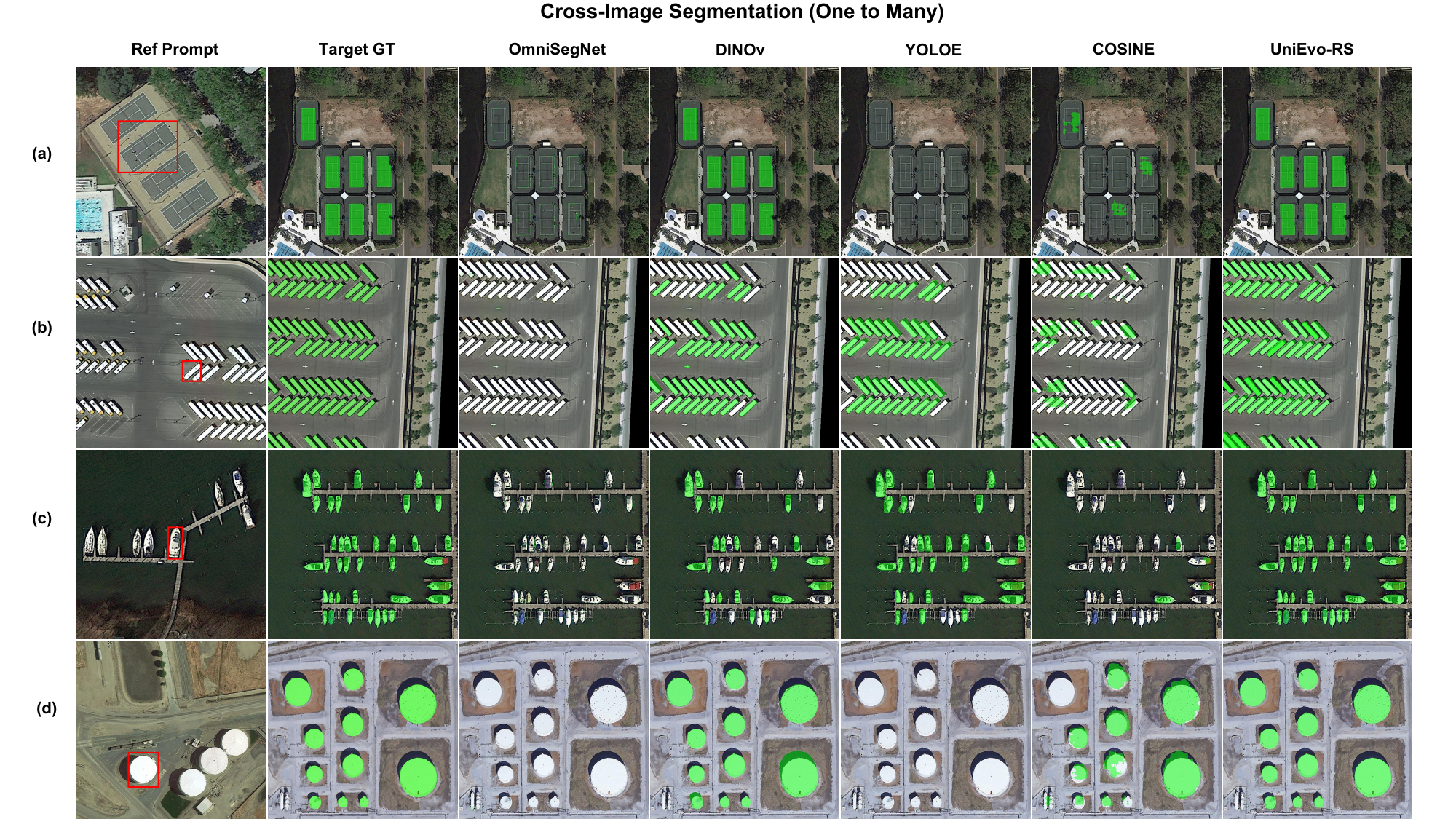}
    \caption{Qualitative comparisons on cross-image one-to-many
    segmentation. A reference instance enclosed by the red bounding box
    is selected from a source image, while all semantically corresponding
    instances must be segmented in a different target image. Green overlays
    denote the target ground-truth or predicted masks.}
    \label{fig:cross_image_qualitative}
\end{figure*}

Figure~\ref{fig:cross_image_qualitative} further evaluates semantic transfer
across different images. Compared with intra-image prompting, this setting is
more challenging because the reference and target instances may differ
substantially in appearance, scale, orientation, density, spatial arrangement,
and imaging context.

In the displayed examples, OmniSegNet identifies only a subset of the
corresponding targets. DINOv and YOLOE recover more matching instances but
still exhibit missed objects or incomplete masks. COSINE produces meaningful
responses in several cases but also introduces fragmented predictions and
false-positive regions.

UniEvo-RS more consistently transfers the region-level semantics of the
reference prompt to the target image. It recovers a larger proportion of
matching instances for densely parked vehicles, differently oriented ships,
multi-scale storage tanks, and repeated sports-field structures. These
results support the effectiveness of the unified visual-prompt interface for
category-level correspondence rather than local appearance matching alone.

\subsubsection{Qualitative Capability Study for Many-to-Many Segmentation}
\label{sec:supp_many_to_many}

The unified visual-prompt interface also supports multi-region prompting.
In the many-to-many setting, multiple reference regions are provided
simultaneously, potentially corresponding to different semantic categories,
and the model is required to segment all target regions associated with the
active prompts.

\begin{figure*}[t]
    \centering
    \includegraphics[width=0.98\textwidth]{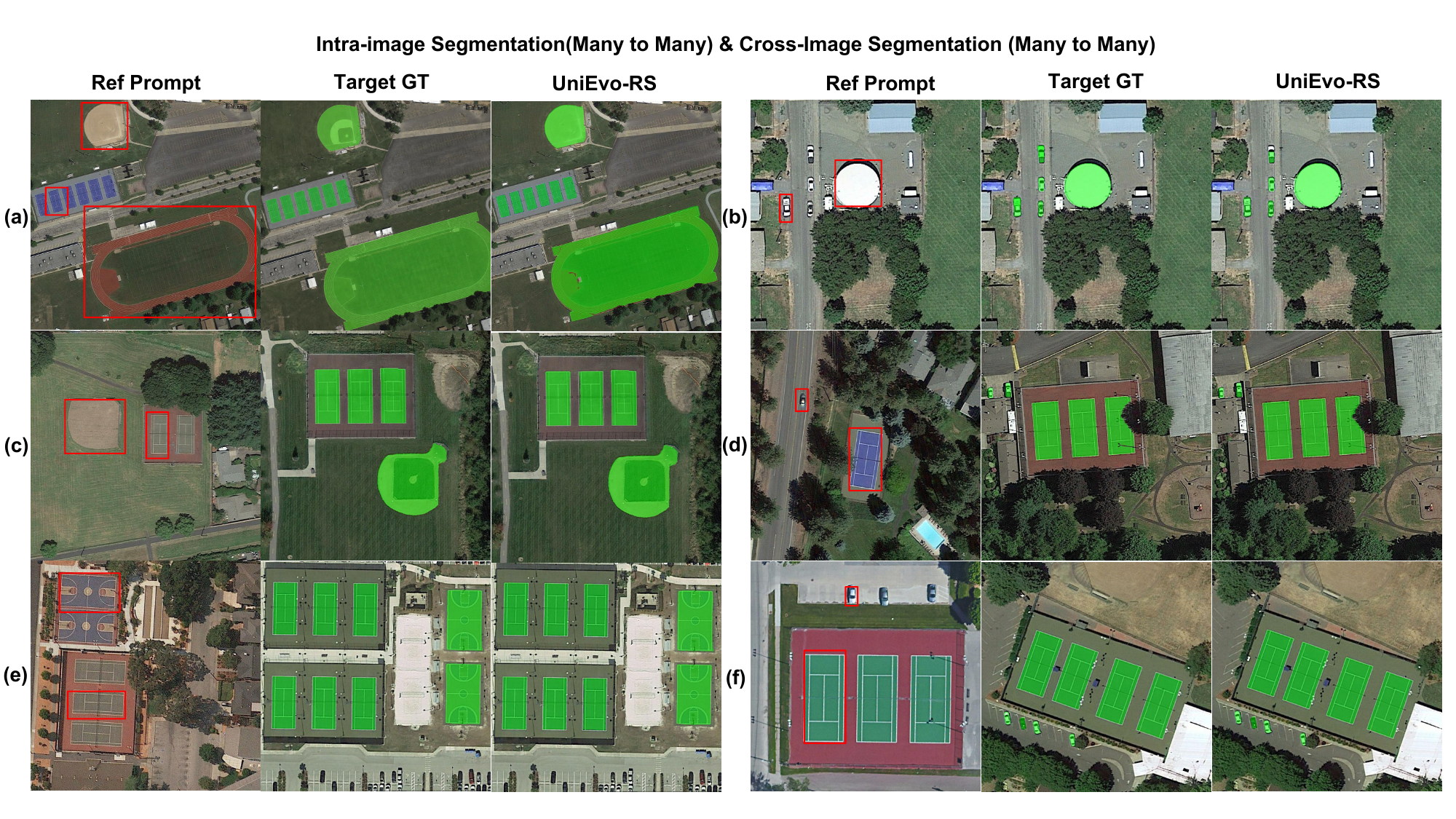}
    \caption{Qualitative demonstrations of many-to-many visual-prompted
segmentation with UniEvo-RS. Panels (a) and (b) show intra-image
many-to-many segmentation, where multiple reference prompts and their
corresponding targets appear in the same image. Panels (c)--(f) show
cross-image many-to-many segmentation, where multiple prompts are selected
from a source image and their semantically corresponding regions are
segmented in a different target image. Red boxes denote the prompted
reference regions, while green overlays visualize the union of the
corresponding target masks. }
    \label{fig:many_to_many_qualitative}
\end{figure*}

As shown in Figure~\ref{fig:many_to_many_qualitative}, UniEvo-RS can jointly
process multiple visual prompts and recover multiple corresponding target
regions within a single inference process. In the intra-image examples, the
model segments regions associated with several active prompts in the same
scene. In the cross-image examples, it transfers the semantics of multiple
prompted regions to a different image despite changes in appearance, scale,
layout, and background context.These examples indicate that the omni-prompt design is not structurally restricted to one-to-many interaction.

\subsection{Representative Exemplar-Driven Prototype Evolution}
\label{sec:supp_prototype_evolution}

\subsubsection{Sequential Batch-Annotation Protocol}
\label{sec:supp_evolution_protocol}

The prototype evolution mechanism is evaluated under a sequential
human-in-the-loop batch-annotation protocol. Representative exemplars refer
to the small subset of samples inspected and corrected by the annotator
before the remaining image collection is processed. UniEvo-RS does not
require a learned exemplar-selection module.

For the $t$-th image in an annotation sequence, prediction uses only the
prototype memory accumulated from previously verified samples:
\begin{equation}
\hat{Y}_{t}
=
\mathcal{F}
\left(
I_{t},
\mathcal{I}_{t},
P_{t};
\mathcal{M}_{t-1}
\right),
\label{eq:supp_sequential_prediction}
\end{equation}
where $I_t$ is the current target image, $\mathcal{I}_t$ is the task
instruction, $P_t$ denotes the active textual or visual prompt, and
$\mathcal{M}_{t-1}$ contains only historical prototype memory.

The annotation or corrected mask $Y_t^{v}$ is revealed only after
$\hat{Y}_t$ has been produced. The comparison between $\hat{Y}_t$ and
$Y_t^{v}$ yields TP, FN, and FP regions, which are then used to update
the memory:
\begin{equation}
\mathcal{M}_{t}
=
\operatorname{Update}
\left(
\mathcal{M}_{t-1},
\operatorname{Decompose}
\left(
\hat{Y}_{t},Y_{t}^{v}
\right)
\right).
\label{eq:supp_sequential_update}
\end{equation}

Therefore, the annotation of the current target image never contributes
to its own prediction. In our experiments, human verification is simulated
by revealing the dataset annotation only after the current prediction has
been generated. This sequential protocol prevents target-mask leakage and
matches the intended practical workflow.

TP and FN features form positive target memory, while FP features form
negative background memory. K-Means independently compresses the three
historical banks into a fixed number of prototype centers. Only these
centers are retained after each update; the complete annotation history is
not cached or injected into subsequent images.

In the following figures, \textit{Static Pred} denotes inference using the
current task instruction and prompt without historical prototype memory.
\textit{Evolved Pred} uses the same current input together with the memory
accumulated from previously verified representative exemplars.

\subsubsection{Generalization to Unseen Instances of Seen Categories}
\label{sec:supp_seen_category_evolution}

In this setting, the evaluated semantic categories are included in
task-specific model training, while the target images and object instances
are unseen. The representative exemplars used to construct the historical
memory are separated from the target images on which the qualitative results
are reported.

\begin{figure*}[t]
    \centering
    \includegraphics[width=0.98\textwidth]{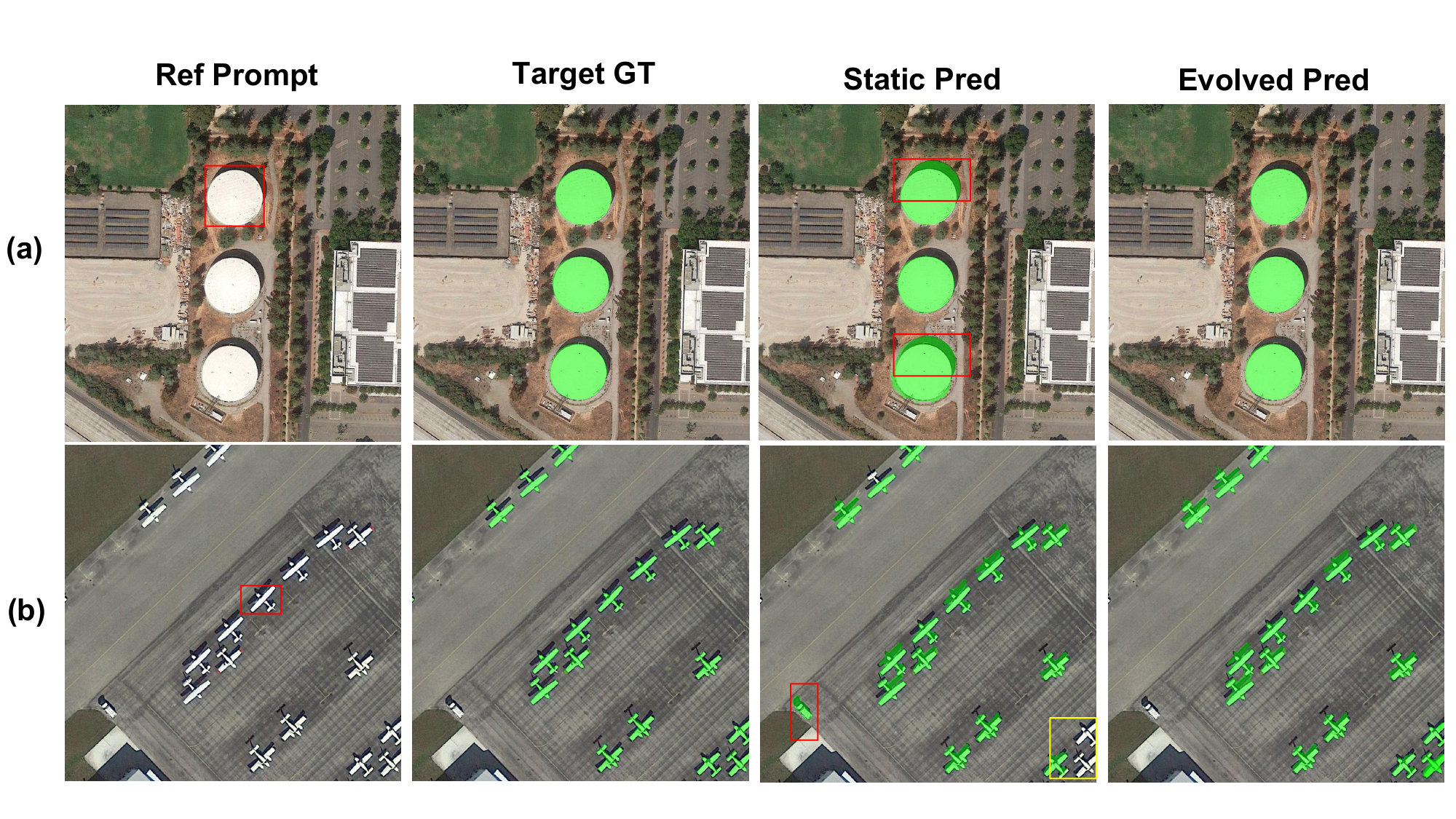}
    \caption{Qualitative comparison of representative exemplar-driven
    prototype evolution on unseen instances of seen categories. From
    left to right, the columns show the current reference prompt, target
    ground truth, static prediction without historical memory, and
    prediction with evolved prototype memory. The displayed categories
    are Storage Tank and Aircraft. Green overlays denote target or
    predicted masks, while colored boxes highlight representative errors
    in the static predictions.In the static predictions, red boxes highlight falsely predicted target pixels, corresponding to false-positive regions, whereas yellow boxes indicate missed target pixels, corresponding to false-negative regions.}
    \label{fig:in_domain_evolution}
\end{figure*}

As shown in Figure~\ref{fig:in_domain_evolution}, static inference already
captures the coarse semantics of Storage Tank and Aircraft but still produces
incomplete target boundaries, missed regions, and responses to visually
confusing structures.

After representative feedback is accumulated, the evolved predictions
provide more complete storage-tank masks and more consistent aircraft
coverage. TP prototypes preserve stable target characteristics, FN prototypes
provide additional evidence for difficult or previously missed target
patterns, and FP prototypes suppress recurrent distractors identified from
earlier verified samples.

These examples indicate that the clustering-compressed prototype memory is
not tied to individual representative samples. Instead, it provides reusable
target and background priors for subsequent unseen images and instances of
the corresponding semantic category.

\subsubsection{Training-Free Adaptation to Held-Out Categories}
\label{sec:supp_unseen_category_evolution}

We further evaluate whether representative feedback can improve batch
annotation for categories excluded from task-specific model training. In
this controlled setting, the model is trained on the original task data
containing categories such as Storage Tank and Aircraft, whereas Wind
Turbine and Vehicle are treated as held-out categories.

At test time, a small number of representative samples from each held-out
category are first predicted and subsequently verified. Their TP, FN, and
FP regions are used to construct category-relevant prototype memory. The
resulting memory is then applied to the remaining images of the same
held-out category without gradient-based fine-tuning or parameter updates.

\begin{figure*}[t]
    \centering
    \includegraphics[width=0.98\textwidth]{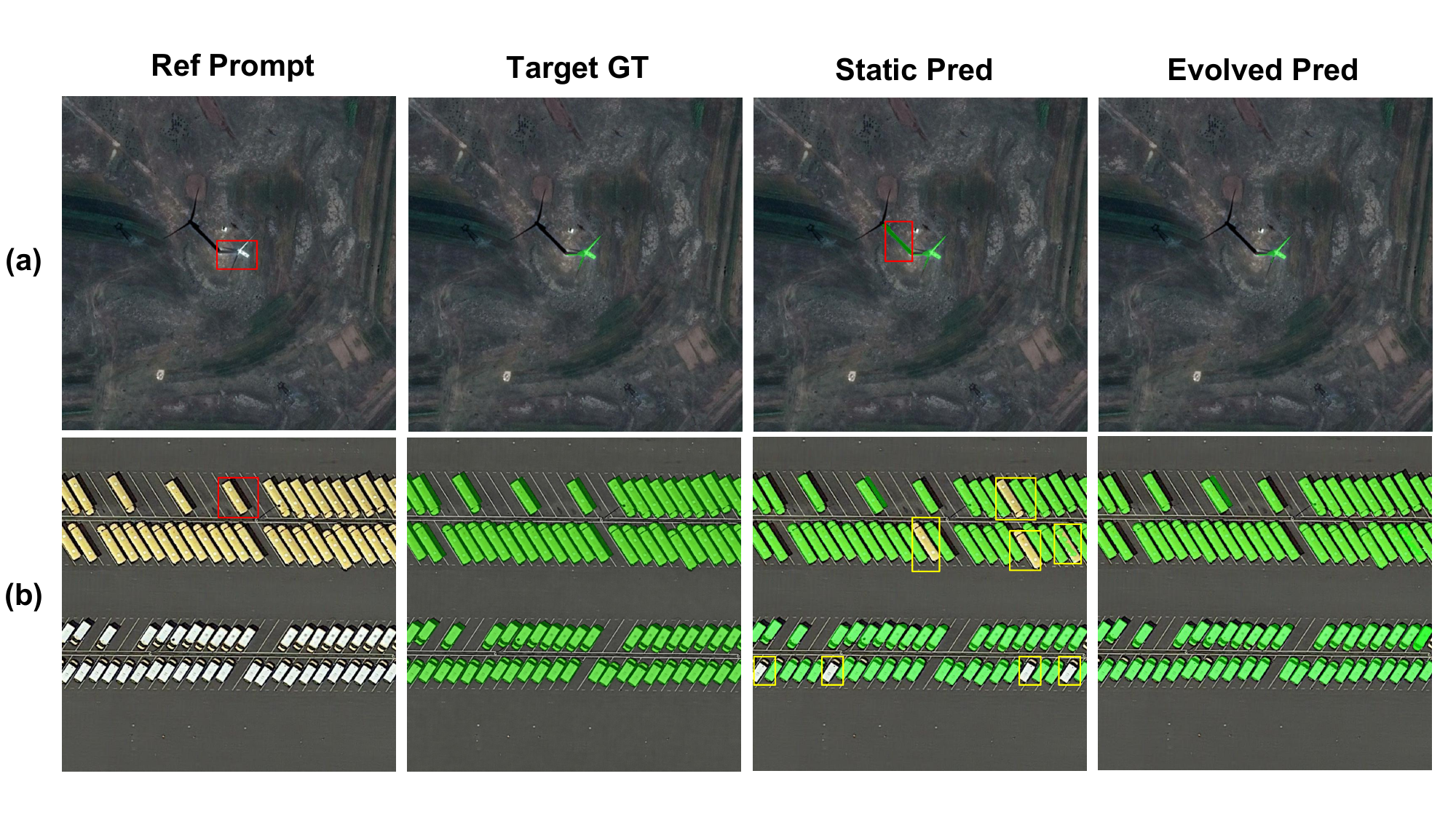}
    \caption{Qualitative comparison of training-free prototype evolution
    on categories held out from task-specific model training. For each
    held-out category, representative samples are first verified to
    construct the historical prototype memory, which is then applied to
    the remaining target images without updating model parameters. From
    left to right, the columns show the current reference prompt, target
    ground truth, static prediction, and evolved prediction. The displayed
    categories are Wind Turbine and Vehicle.In the static predictions, red boxes highlight falsely predicted target pixels, corresponding to false-positive regions, whereas yellow boxes indicate missed target pixels, corresponding to false-negative regions.}
    \label{fig:unseen_category_evolution}
\end{figure*}

Figure~\ref{fig:unseen_category_evolution} shows that the static model can
obtain initial responses for held-out categories through the current visual
prompt, but still suffers from missed regions and recurrent false positives.
After prototype evolution, the positive memory improves target recovery,
while the negative memory suppresses several background responses that
recur across the annotation batch.



\subsubsection{Positive Prototype Allocation}
\label{sec:prototype_allocation}

The positive memory contains both TP and FN prototypes. TP prototypes preserve
stable characteristics of correctly recognized targets, whereas FN prototypes
capture difficult patterns associated with targets missed by static inference.
We keep the total positive-memory capacity fixed at eight and vary its
allocation between TP and FN prototypes.

\begin{table}[t]
\centering
\small
\setlength{\tabcolsep}{4.0pt}
\renewcommand{\arraystretch}{1.15}

\textbf{(a) Storage Tank}
\par\vspace{1mm}

\begin{tabular*}{\columnwidth}{
@{\extracolsep{\fill}}cccc@{}
}
\toprule
\textbf{Exp.}
& \textbf{TP:FN}
& \textbf{gIoU}
& \textbf{mIoU} \\
\midrule
1 & 1:7 & \textbf{56.54} & 75.87 \\
2 & 2:6 & 56.28 & \textbf{76.30} \\
3 & 4:4 & 40.95 & 67.52 \\
4 & 6:2 & 54.53 & 71.99 \\
5 & 7:1 & 54.42 & 71.90 \\
\bottomrule
\end{tabular*}

\vspace{3mm}

\textbf{(b) Vehicle}
\par\vspace{1mm}

\begin{tabular*}{\columnwidth}{
@{\extracolsep{\fill}}cccc@{}
}
\toprule
\textbf{Exp.}
& \textbf{TP:FN}
& \textbf{gIoU}
& \textbf{mIoU} \\
\midrule
1 & 1:7 & 40.20 & 59.29 \\
2 & 2:6 & \textbf{41.24} & \textbf{59.50} \\
3 & 4:4 & 40.39 & 58.61 \\
4 & 6:2 & 40.17 & 59.35 \\
5 & 7:1 & 32.25 & 57.21 \\
\bottomrule
\end{tabular*}

\caption{Ablation study of different TP-to-FN prototype allocations
under a fixed positive-memory budget of eight. The best result for each
metric is shown in bold.}
\label{tab:tp_fn_ratio}
\end{table}

As shown in Table~\ref{tab:tp_fn_ratio}, FN-dominant allocations are generally
more effective for the two evaluated challenging categories. For Storage
Tank, the 1:7 allocation obtains the highest gIoU, whereas 2:6 obtains the
highest mIoU. For Vehicle, 2:6 performs best on both metrics.

The results indicate that allocating sufficient capacity to FN prototypes is
important because they preserve hard target patterns that are absent from
static predictions. We use $m=2$ TP prototypes and $n=6$ FN prototypes in
the remaining experiments because this setting provides the most consistent
overall behavior across the two evaluated categories.

\subsubsection{Hyperparameter Sensitivity}
\label{sec:sensitivity_analysis}

We evaluate the positive-prior injection strength $\alpha$ and the negative
spatial suppression strength $\lambda$. The two parameter studies are
conducted sequentially. We first vary $\alpha$ while retaining the original
suppression coefficient $\lambda=5.0$. After selecting $\alpha=1.0$, we fix
$\alpha$ and refine $\lambda$ within a narrower range. Because the fixed
value of $\lambda$ differs between the two blocks, their absolute metric
values should not be directly compared.

\begin{table}[t]
\centering
\small
\setlength{\tabcolsep}{8pt}
\renewcommand{\arraystretch}{1.08}
\begin{tabular*}{\columnwidth}{
@{\extracolsep{\fill}}cccc@{}
}
\toprule
\textbf{Parameter}
& \textbf{Value}
& \textbf{gIoU}
& \textbf{mIoU} \\
\midrule

\multicolumn{4}{l}{
\textit{Positive prototype injection strength}
$\alpha$ \textit{ ($\lambda=5.0$)}
} \\
\midrule
$\alpha$ & 0.8 & 49.54 & 74.66 \\
$\alpha$ & 0.9 & 48.56 & 73.44 \\
$\alpha$ & 1.0 & \textbf{49.83} & \textbf{74.91} \\
$\alpha$ & 1.1 & 49.38 & 73.95 \\

\midrule

\multicolumn{4}{l}{
\textit{Negative suppression strength}
$\lambda$ \textit{ ($\alpha=1.0$)}
} \\
\midrule
$\lambda$ & 0.8 & 51.89 & 78.18 \\
$\lambda$ & 0.9 & 51.91 & \textbf{78.20} \\
$\lambda$ & 1.0 & 51.92 & \textbf{78.20} \\
$\lambda$ & 1.1 & \textbf{51.94} & 78.19 \\
\bottomrule
\end{tabular*}
\caption{Sensitivity analysis of the positive prototype injection
strength $\alpha$ and negative spatial suppression strength $\lambda$.}
\label{tab:sensitivity}
\end{table}

As shown in Table~\ref{tab:sensitivity}, $\alpha=1.0$ provides the best
gIoU and mIoU in the positive-guidance study. With $\alpha$ fixed to 1.0,
the results remain stable when $\lambda$ varies from 0.8 to 1.1.
We choose $\lambda=1.1$ because it achieves the highest gIoU while
maintaining an mIoU comparable to the best-performing settings.

The relatively narrow variation within each controlled block indicates that
prototype evolution does not depend on a highly specific parameter value.
Unless otherwise stated, all remaining experiments use
$\alpha=1.0$ and $\lambda=1.1$.

\subsubsection{Fixed-Budget Performance--Efficiency Analysis}
\label{sec:efficiency_analysis}

A central design objective of prototype evolution is to reuse historical feedback without causing unbounded memory growth as more samples are verified. We therefore evaluate two factors: the number of retained positive and negative prototypes and the annotation-sequence length.

\begin{table}[t]
\centering
\small
\setlength{\tabcolsep}{3pt}
\renewcommand{\arraystretch}{1.08}
\begin{tabular*}{\columnwidth}{
@{\extracolsep{\fill}}ccccc@{}
}
\toprule
\textbf{Pos./Neg.}
& \textbf{Mem.}
& \textbf{Lat.}
& \textbf{FPS}
& \textbf{mIoU} \\
\midrule
4/4   & 8.34 & 410.44 & 2.44 & 77.29 \\
8/8   & 8.34 & 408.61 & 2.45 & 78.19 \\
16/16 & 8.34 & 391.82 & 2.55 & 77.85 \\
32/32 & 8.34 & 392.14 & 2.55 & 77.37 \\
\bottomrule
\end{tabular*}
\caption{Performance--efficiency analysis under different positive and
negative prototype-memory budgets. Memory and latency are reported in GB
and ms, respectively.}
\label{tab:efficiency_budget}
\end{table}

Table~\ref{tab:efficiency_budget} varies the positive and negative prototype
budgets while keeping the annotation sequence fixed. The 8/8 configuration
achieves the highest mIoU. Increasing the retained budget to 16/16 or 32/32
does not provide further accuracy gains, suggesting that retaining too many prototypes may cause the memory to overfit instance-specific details rather than capturing generalizable category-level semantics, thereby hindering cross-image generalization.

Peak GPU memory remains virtually unaffected by the prototype budget, as the prototype tensors introduce minimal overhead compared to the heavy memory requirements of the visual encoder, LLM, and mask decoder.
Latency and FPS remain within a similar range and do not exhibit systematic
growth as the retained prototype budget increases.

\begin{table}[t]
\centering
\small
\setlength{\tabcolsep}{3pt}
\renewcommand{\arraystretch}{1.08}
\begin{tabular*}{\columnwidth}{
@{\extracolsep{\fill}}ccccc@{}
}
\toprule
\textbf{Seq. Len.}
& \textbf{Mem.}
& \textbf{Lat.}
& \textbf{FPS}
& \textbf{mIoU} \\
\midrule
50   & 7.79 & 424.72 & 2.35 & 77.06 \\
100  & 7.79 & 402.88 & 2.48 & 76.67 \\
500  & 8.34 & 399.20 & 2.51 & 77.71 \\
1000 & 8.34 & 431.95 & 2.32 & 77.12 \\
\bottomrule
\end{tabular*}
\caption{Performance--efficiency analysis under different annotation
sequence lengths. Memory and latency are reported in GB and ms,
respectively.}
\label{tab:efficiency_sequence}
\end{table}

Table~\ref{tab:efficiency_sequence} fixes the retained prototype budget and
increases the annotation-sequence length from 50 to 1000. Across this
twenty-fold increase, memory usage remains within a narrow range and latency
does not increase with sequence length.

The memory increases from 7.79 GB to 8.34 GB when the prototype banks reach
their configured capacity, after which the model-facing memory remains
bounded. This behavior is consistent with the fixed-budget update mechanism:
newly verified region embeddings are temporarily combined with the previously
retained centers, compressed through K-Means, and discarded after the updated
centers have been obtained. Consequently, the model receives a bounded number
of prototype tokens rather than the complete annotation history.

\begin{table}[!ht]
\centering
\small
\setlength{\tabcolsep}{3.5pt}
\renewcommand{\arraystretch}{1.12}
\begin{tabularx}{\columnwidth}{@{}c>{\centering\arraybackslash}Xcc@{}}
\toprule
\textbf{Exp.} & \textbf{Compression Strategy} & \textbf{gIoU} & \textbf{mIoU} \\
\midrule
1 & Random sampling & 39.87 & 58.28 \\
2 & Importance-based selection & 18.46 & 54.56 \\
3 & Similarity-based merging & 31.28 & 57.06 \\
4 & K-Means clustering & \textbf{41.24} & \textbf{59.50} \\
\bottomrule
\end{tabularx}
\caption{Comparison of memory compression strategies under a fixed budget. Positive memory uses eight prototypes (two TP and six FN), while negative memory uses eight FP prototypes.}
\label{tab:compression_strategy}
\vspace{-13pt}
\end{table}

\subsection{Additional Implementation Details}
\label{sec:supp_implementation}

\subsubsection{Model Configuration}
\label{sec:supp_model_configuration}

Following PSALM, UniEvo-RS is initialized from its pretrained weights and
uses a Phi-based LLM, a Swin-B visual encoder, an MSDeformAttn pixel decoder,
and a Mask2Former-style predictor. The mask predictor has a hidden dimension
of 256, 100 object queries, 8 attention heads, and 9 transformer decoder
layers, resulting in 10 prediction stages when the initial query prediction
is included.

The visual encoder is frozen during training. The multimodal projector, LLM,
pixel decoder, and mask predictor are fine-tuned jointly. Visual prompts are
represented as points, bounding boxes, scribbles, or region masks.

\subsubsection{Multi-Task Training Configuration}
\label{sec:supp_training_configuration}

We train UniEvo-RS for 10 epochs using a base learning rate of
$1\times10^{-6}$, a cosine learning-rate schedule, and a warmup ratio of
0.03. Batch-wise alternating multi-task training is used: each mini-batch
is sampled from one task setting, and the task source alternates across
successive mini-batches.

The per-device batch size is 4, and the maximum multimodal sequence length
is 2048. The same shared architecture and mask-generation pathway are used
for all five quantitative task settings.

\subsubsection{Prototype-Evolution Configuration}
\label{sec:supp_prototype_configuration}

The positive memory contains $m=2$ TP prototypes
and $n=6$ FN prototypes, while the negative memory contains $r=8$ FP
prototypes. The TP, FN, and FP update pools are clustered independently.

Whenever newly verified region embeddings are added, the previously retained
centers and the new embeddings are temporarily combined and K-Means is rerun.
Only the updated cluster centers are retained after each evolution step.
The full historical feature pool is discarded.

We use a positive-prior injection strength of $\alpha=1.0$, a negative
confidence threshold of $\tau=0.7$, and a spatial suppression strength of
$\lambda=1.1$. Prototype guidance is activated only during representative
exemplar-driven batch inference, and all model parameters remain fixed.

\subsection{Memory Compression Strategy}
\label{sec:memory_compression}

Table~\ref{tab:compression_strategy} compares four strategies under the same
fixed memory budget. K-Means clustering performs best on both metrics,
suggesting that preserving representative feature modes is more effective than
sample-level selection or merging in this setting. This result supports its use
for compressing the evolving prototype memory while keeping the model-facing
memory size fixed.


\end{document}